\documentclass[conference]{IEEEtran}
\IEEEoverridecommandlockouts
\usepackage{cite}
\usepackage{amsmath,amssymb,amsfonts}
\usepackage{algorithmic}
\usepackage{graphicx}
\usepackage{textcomp}
\usepackage{xcolor}
\def\BibTeX{{\rm B\kern-.05em{\sc i\kern-.025em b}\kern-.08em
    T\kern-.1667em\lower.7ex\hbox{E}\kern-.125emX}}
\begin{document}

\title{User, Robot, Deployer: A New Model for Measuring Trust in HRI}
\author{\IEEEauthorblockN{1\textsuperscript{st} David Cameron}
\IEEEauthorblockA{\textit{Information School} \\
\textit{The University of Sheffield}\\
Sheffield, UK. \\
d.s.cameron@sheffield.ac.uk \\
ORCID 0000-0001-8923-5591}
\and
\IEEEauthorblockN{2\textsuperscript{nd} Emily C. Collins}
\IEEEauthorblockA{\textit{Department of Computer Science} \\
\textit{University of Manchester}\\
Manchester, UK. \\
e.c.collins@manchester.ac.uk \\
ORCID 0000-0001-9396-536X}
}
\maketitle

\begin{abstract}
There is an increasing interest in considering, implementing, and measuring trust in human-robot interaction (HRI). Typically, this centres on influencing user trust within the framing of HRI as a dyadic interaction between robot and user. We propose this misses a key complexity: a robot's trustworthiness may also be contingent on the user's relationship with, and opinion of, the individual or organisation deploying the robot. Our new HRI triad model (User, Robot, Deployer), offers novel predictions for considering and measuring trust more completely.
\end{abstract}


\section{Introduction}
Research aimed at measuring trust in Human-Robot Interaction (HRI) scenarios is fundamental to the development of readily adoptable robotics and autonomous systems (RAS) technologies. Both social and industrial RAS need to be presented to users as approachable, efficient, and safe. Existing research commonly explores trustworthy HRI via a dyadic approach: labelling an individual user as one half of an interaction, and attempting to optimise a trust benchmark outcome from said individual both during and after an interaction with a robotic agent \cite{yagoda2012you,hancock2011meta,schaefer2016meta}. Indeed, there are a range of identified factors - from the robot itself, the user, and the environment - that may affect trust measurement in HRI \cite{hancock2011meta}. For example, the use of social/social-like behaviour expressed by a robot is considered as a factor against which a user's trust in that particular robot is measured \cite{law2021trust}. However, in this position paper we argue that this range of factors is incomplete. Borrowing from Vicente's model of the Tech Ladder \cite{vicente2004human}, we propose that measuring trust in HRI requires a triadic approach, in which the deployer of the robot, and the relationship said deployer has with the robot-user, is also considered as an impacting factor.

\subsection{HRI across multiple levels}
Vicente writes that technology is \emph{any} human invention to meet a human or societal need \cite{vicente2004human}. Where RAS and Powerplants are examples of hard (i.e. tangible) technology, other inventions such as Staff-Workload-Allocation and Democracy are conversely soft technologies. Interactions with deployed robotics can be studied at varying levels of abstraction (the Physical, Psychological, Team, Organisational, and Political levels), which all connect. Given this, we propose that whilst the lower levels of Physical, Psychological and potentially even Team are relatively well-studied in understanding trust in HRI, the under-considered higher levels of Organisational and Political also impact trust in even small-scale interactions.

\section{The Social Triad of HRI}

\begin{figure}[ht]
  \includegraphics[width=\columnwidth]{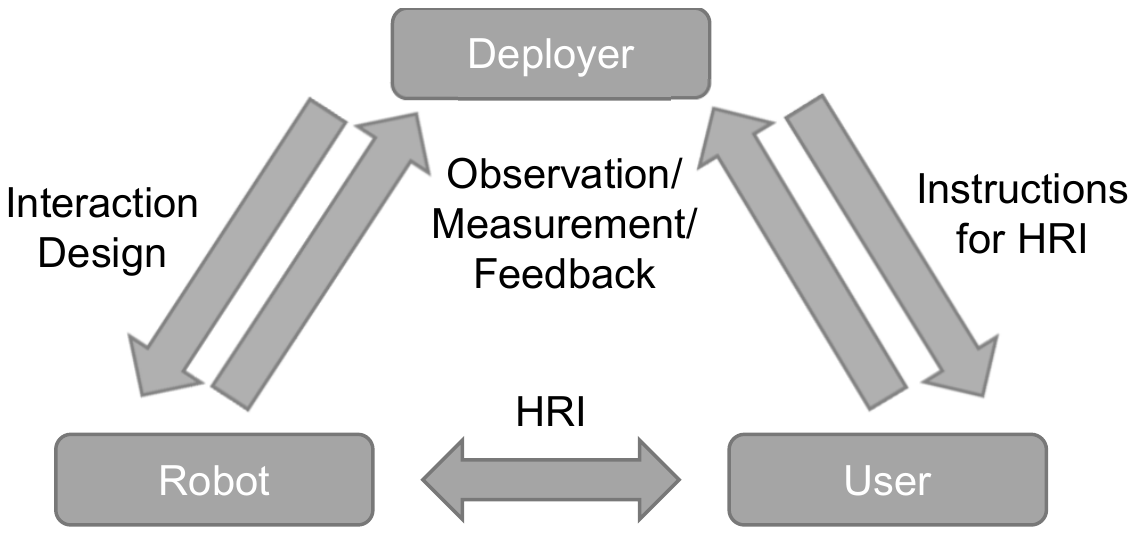}
  \caption{Pathways of communication for an HRI scenario}
  \label{fig:connect}
\end{figure}

While the commonly-held approach to HRI as interaction between user and robot contained within a specified scenario may reliably capture key elements of interest, this view marginalises the influence higher levels, crucially the person/agent deploying the robot to the user, have on the `shape' of the scenario and the interaction experience. Where a programmer, owner, or researcher using a robot for HRI may consider themselves to be external to the scenario, their influence may still be experienced by the user (the `H' in HRI). Our model (Fig. \ref{fig:connect}) seeks to explicitly include the role that these `external' agents may have on HRI, and their impact on measurements of trust in the interaction.

Building on the approach to trust in HRI from a social perspective, we expand the common view of HRI as a social dyad between User and Robot to view HRI as a social triad that includes the Deployer. `Robot' and `User' as terms should be familiar to a researcher in HRI; for this model, `Deployer' is the interested external agent, such as a researcher or manager, and/or body, such as a corporation, who brought the scenario into being. The User's perceived relationship with the Deployer - both any literal relationship and their regard for the deployer as representing the Organisational and Political decisions in creating the scenario - may influence their views towards the scenario and interactions with the robot.

\subsection{Interactions in the Triad}

Perhaps the most apparent interaction for study in the social triad is the (often reciprocal) one between the User and the Robot (i.e. HRI). A user may influence a robot's actions through their own behaviour, either directly controlling or otherwise affecting the robot. In return, users' experiences and their evaluations of the interaction may be influenced by social-like behaviours of a robot. Specifically, the trust an individual has towards a robot may be affected by the robot's appearance and behaviours \cite{hancock2011meta} and social strategies to regulate trust \cite{cameron2021effect,geiskkovitchchildren,mota2016playing}.

The interaction from Deployer (D) to Robot (R) includes the production of robot behaviours through direct control of a robot (e.g., scenarios requiring Wizard of Oz control \cite{riek2012wizard}), specifying goals for the robot, or developing architectures to generate behaviour \cite{trafton2013act}. While the Deployer may be responsible for programming the robot, it is not necessary for them to be present to have influence in the social triad. The Deployer specifies the contexts for which robots may be used and the bounds for a robot's interaction with the User (U)\cite{cameron2015framing}.

D to U: Like the Deployer determines the bounds of a Robot's behaviour, their construction of an interaction scenario may constrain the User. Communication from the Deployer and existing or emerging relationships between Deployer and User may shape a User's approach to HRI: Users might evaluate a scenario based on views of the Deployer's behaviours and intentions while ostensibly evaluating the robot.

U to D: feedback on the scenario - either passively when under observation or actively through agreed feedback channels (in research, this may be questionnaires, interviews etc.; in industry, this may be performance appraisals etc.).

R to D: Sharing of information on the interaction (passively through the behaviour) or actively through recorded metrics from the interaction, potentially of both Robot and User.

Thus, where Users may need to trust Robots, they also need to trust Deployers (e.g., Deployers will not allow the User to come to harm (physical or psychological); they have programmed the Robot to operate appropriately (and transparently); they have specified an HRI scenario which complies with necessary safety/ethical regulations; they do not use information gathered from HRI to adversely affect Users.)

\subsection{Interaction Scenarios}
What the model offers: HRI is a scenario into which a user has entered but not-necessarily specified. Trust in HRI may depend on the communication and trust towards the deployer as well as trust towards the robot; any measurement of trust the user may have of the robot may also reflect how much trust the user has in the deployer. Further, the model can be used to highlight the importance of transparency in understanding trustworthy HRI. As well as knowing the processes involved in an HRI scenario, a user also wants to know if their interaction is going to be reported (e.g., is the robot serving as a secondary channel for recording the user beyond any direct interaction from U to D?). Although a user can withdraw from an HRI scenario, given the role of D, they may have relatively little power to shape or define the HRI scenario itself. Thus, crucially, the user, at the whim of the deployer, may provide a trust assessment of the robot that is biased due to the influence of the deployer.

\section{Proposed Directions for Future Study}
In sum, our model predicts: 
\begin{enumerate}
    \item The perceived trustworthiness of the deployer is positively associated with trust in interactions.
    \item Participant's articulation of trustworthiness that includes views towards the deployer would explain some variation in user-differences in interactions that are otherwise attributed to various factors towards technology, such as experience with robotics and demographics.
    \item Attempts to scaffold trust through enacting social/social-like behaviours in the robot may even have a counter-intuitive negative impact on perceived trustworthiness in scenarios where users' trust towards the deployer is low.
    \item Inclusion of the user in HRI scenario development promotes trust in the deployer and, indirectly, the robot.
    \end{enumerate}
We hope that future work exploring methods and metrics to measure trust in HRI takes this model into consideration.

\section*{ACKNOWLEDGMENT}
This work was funded by UKRI Trustworthy Autonomous Systems (TAS) programme, through the UKRI Strategic Priorities Fund and delivered by the Engineering and Physical Sciences Research Council (EPSRC).
\bibliography{IEEEabrv}
\end{document}